\documentclass[conference]{IEEEtran}

\usepackage{cite}
\usepackage{amsmath,amssymb,amsfonts}
\usepackage{algorithmic}
\usepackage{graphicx}
\usepackage{textcomp}
\usepackage[table]{xcolor}
\usepackage{tikz}
\usepackage{hyperref}
\usepackage{caption}
\usepackage{booktabs}
\usepackage{multirow}
\usepackage{pgfplots}
\usepackage{times}
\usepackage{orcidlink}
\pgfplotsset{compat=newest}

\def\BibTeX{{\rm B\kern-.05em{\sc i\kern-.025em b}\kern-.08em
    T\kern-.1667em\lower.7ex\hbox{E}\kern-.125emX}}

\newcommand\Tstrut{\rule{0pt}{2.6ex}}
\newcommand\Bstrut{\rule[-0.9ex]{0pt}{0pt}}

\definecolor{lime}{HTML}{A6CE39}
\DeclareRobustCommand{\orcidicon}{
	\begin{tikzpicture}
		\draw[lime, fill=lime] (0,0) 
		circle [radius=0.16] 
		node[white] {{\fontfamily{qag}\selectfont \tiny ID}};
		\draw[white, fill=white] (-0.0625,0.095) 
		circle [radius=0.007];
	\end{tikzpicture}
	\hspace{-2mm}
}

\foreach \x in {A, ..., Z}{\expandafter\xdef\csname orcid\x\endcsname{\noexpand\href{https://orcid.org/\csname orcidauthor\x\endcsname}
		{\noexpand\orcidicon}}
}

\makeatletter
\newcommand{\linebreakand}{%
\end{@IEEEauthorhalign}
\hfill\mbox{}\par
\mbox{}\hfill\begin{@IEEEauthorhalign}
}
\makeatother

\begin{document}

\title{
	LIDSNet: A Lightweight on-device Intent Detection model using Deep Siamese Network
}

\author{
	\IEEEauthorblockN{
		Vibhav Agarwal \orcidlink{0000-0002-2029-9885},
		Sudeep Deepak Shivnikar \orcidlink{0000-0002-9586-3540},
		Sourav Ghosh \orcidlink{0000-0003-1866-1408},
		Himanshu Arora,
		Yashwant Saini
	}
	\IEEEauthorblockA{\textit{Samsung R\&D Institute Bangalore}, Karnataka, India 560037\\
		Email: \{
		vibhav.a,
		s.shivnikar,
		sourav.ghosh, 
		him.arora,
		yash.saini
		\}@samsung.com
	}
}

\maketitle

\begin{abstract}

	Intent detection is a crucial task in any Natural Language Understanding (NLU) system and forms the foundation of a task-oriented dialogue system. To build high-quality real-world conversational solutions for edge devices, there is a need for deploying intent detection model on device. This necessitates a light-weight, fast, and accurate model that can perform efficiently in a resource-constrained environment. To this end, we propose LIDSNet, a novel lightweight on-device intent detection model, which accurately predicts the message intent by utilizing a Deep Siamese Network for learning better sentence representations. We use character-level features to enrich the sentence-level representations and empirically demonstrate the advantage of transfer learning by utilizing pre-trained embeddings. Furthermore, to investigate the efficacy of the modules in our architecture, we conduct an ablation study and arrive at our optimal model. Experimental results prove that LIDSNet achieves state-of-the-art competitive accuracy of 98.00\% and 95.97\% on SNIPS \cite{coucke2018snips} and ATIS \cite{price-1990-evaluation} public datasets respectively, with under 0.59M parameters. We further benchmark LIDSNet against fine-tuned BERTs and show that our model is at least 41x lighter and 30x faster during inference than MobileBERT \cite{sun-etal-2020-mobilebert} on Samsung Galaxy S20 device, justifying its efficiency on resource-constrained edge devices.

\end{abstract}


\section{Introduction}\label{sec:introduction}

Identifying message intents from natural language utterances is a crucial task for conversational systems. In applications ranging from natural language response generation to offering intelligent suggestions, understanding the primary intent of the context communication is critical. Traditionally, research on intent detection has focused on this task with the assumption that training and inference would be performed on a well-equipped server or cloud infrastructure. This has led to the unsuitability of existing machine learning approaches for real-world dialog systems in low-resource edge devices due to their high latency and reliance on huge pre-trained models. Lately, there has been an increased academic and commercial interest in supporting AI solutions that can work directly on a user's device on local data \cite{9364648, agarwal-etal-2020-emplite}. On-device AI models have the potential to support intent detection in real-time at low latency and also helps in enhancing the privacy of sensitive user data like smartphone messages.

\begin{figure}[t]
	\centering
	\begin{minipage}[t]{0.58\linewidth}
		\includegraphics[width=\linewidth]{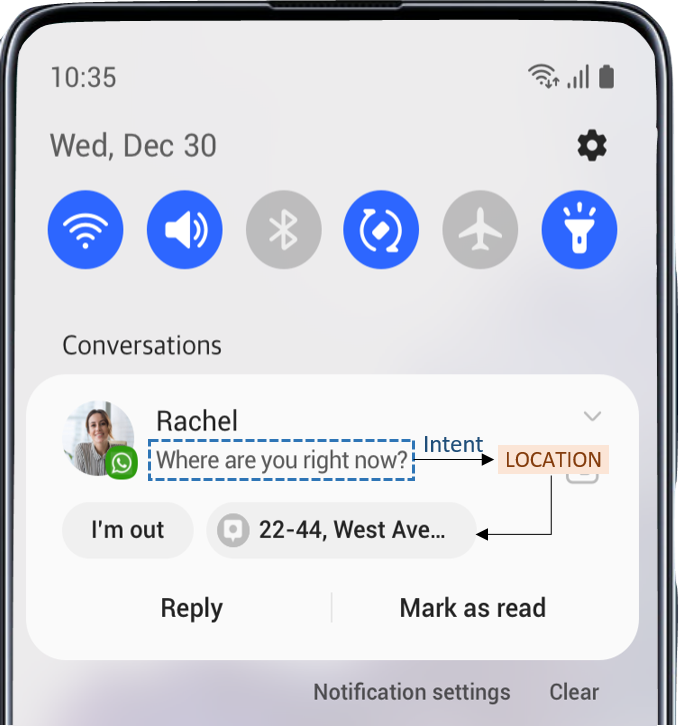}
		\label{fig:teaserPhone}
	\end{minipage}
	\begin{minipage}[t]{0.35\linewidth}
		\includegraphics[width=\linewidth]{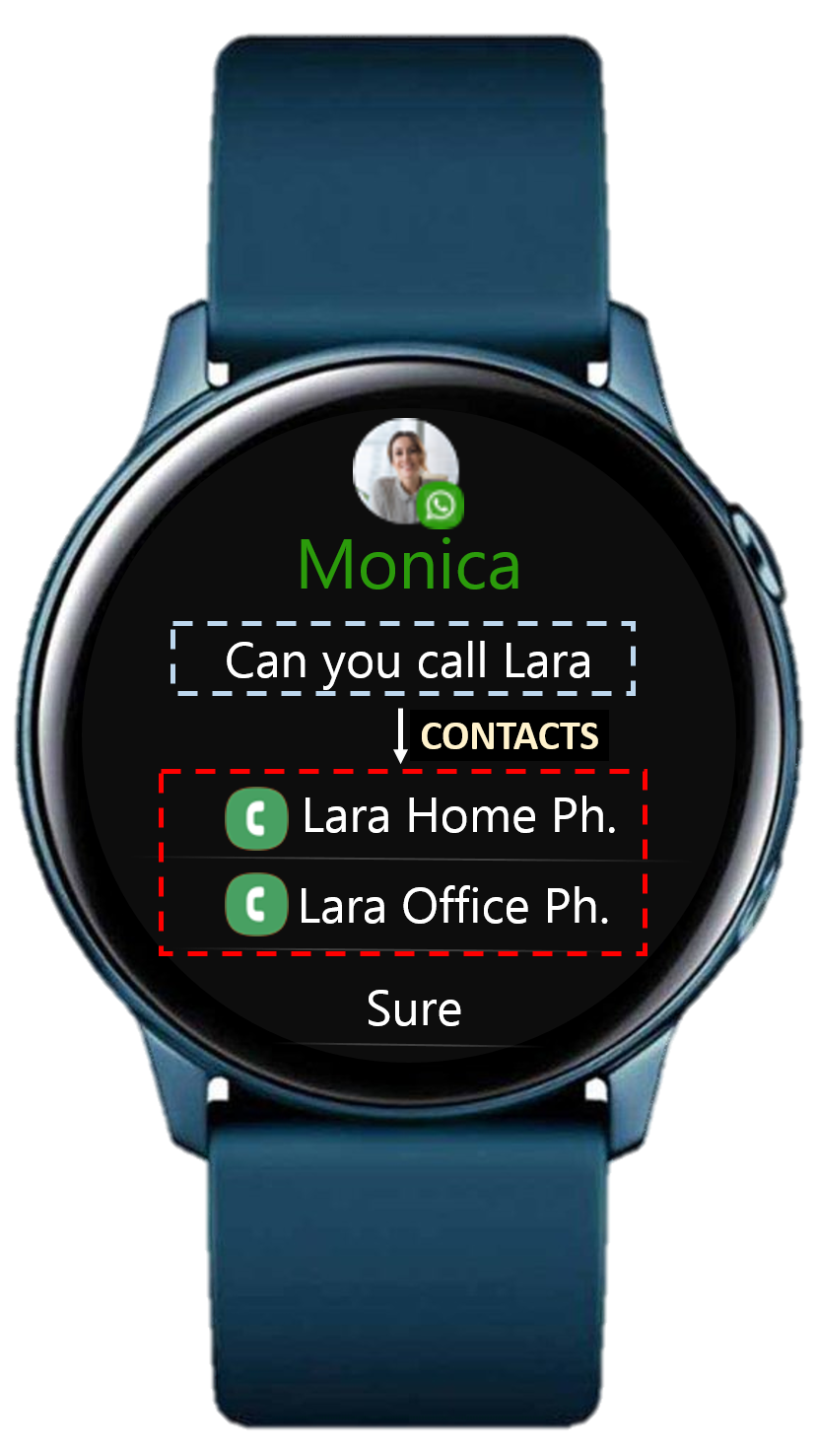}
		\label{fig:teaserWatch}
	\end{minipage}
	\caption{\textbf{On-device Intent Understanding from conversational messages}: {\normalfont A critical problem for downstream tasks like response generation and intelligent suggestion}}
	\label{fig:teaser}
\end{figure}

Recently, Siamese Networks is being popularly used for few-shot learning and similarity tasks. Siamese Networks with Triplet Loss (SN-TL) brings representations of relevant inputs closer in latent space. Huang et al. \cite{Huang_2017} have demonstrated the effectiveness of using SN-TL in the vision domain. Reimers and Gurevych \cite{reimers-gurevych-2019-sentence} modifies pre-trained BERT with SN-TL to derive contextual sentence embeddings. Dhaliwal et al. \cite{9364403} uses a similar approach to bring representations of domain specific synonyms closer and later classify them. This motivates us to leverage Siamese Networks with Triplet Loss for intent detection task. In the context of intent detection, this can bring sentence representations of utterances that belong to the same intent class, closer in latent space. The knowledge gained from this task can be further fine-tuned for classification task. This enables us to take advantage of transfer learning.

In the current work, we propose a novel approach for intent classification with two phase training. In the first phase, we train an encoder using a Siamese Network to learn sentence representations. This functions as a feature extractor that takes into account both character and word level features. This is further fine-tuned for intent classification task in phase II training. Our primary contributions are summarized as follows:

\begin{itemize}
	\item We present a lightweight framework, LIDSNet, to accurately predict message intent as shown in Fig.~\ref{fig:teaser} by incorporating a Siamese Network with Triplet Loss.
	\item We demonstrate the efficacy of using Siamese Network in learning better sentence representations by conducting an ablation study, discussed in subsection \ref{sec:ablationStudy}.
	\item We present promising benchmarking results with LIDSNet against off-the-shelf SOTA models, achieving high accuracy with the lowest memory footprint, as detailed in subsection \ref{sec:comparisonWithSOTA}.\looseness=-1
	\item We benchmark LIDSNet against various variants of BERT in subsection \ref{sec:computationalExperiments}, and empirically demonstrate that LIDSNet outperforms all the baselines and SOTA models in terms of accuracy-size trade-off.
\end{itemize}

Experiments show that LIDSNet achieves an accuracy of 98.00\% and 95.97\% on SNIPS and ATIS datasets respectively. Compared to different versions of BERT, LIDSNet attains 9x-87x speedup in inference time and 24x-186x reduction in the number of model parameters.

\section{Related Work}\label{sec:relatedWork}

Early research on intent detection include Maximum Entropy Markov Models (MEMM) \cite{toutanvoa-manning-2000-enriching}. This task has also been approached using Support Vector Machines (SVM) \cite{5947649}. Lafferty et al. \cite{10.5555/645530.655813} shows that CRF based methods to build probabilistic models for segmentation and labelling sequence data perform better than MEMMs. Following this, Purohit et al. \cite{7463729} demonstrates the effectiveness of using knowledge-guided patterns in short-text intent classification. While such Shallow Learning techniques show decent results on data adhering to a command-driven pattern, their performance in natural language, filled with colloquials and lingos, is lacking.

The emergence of deep learning effectively alleviates the constraints of statistical methods and achieves state-of-the-art (SOTA) results from natural language processing to computer vision. The problem statements explored using deep learning techniques include intent detection and the related problem of slot filling, which aims to extract the values of certain types of attributes for a given entity from the input \cite{qin2019stackpropagation}.

The choice of embeddings plays an important role in the design decision of most intent detection approaches. Instead of using token-level word embeddings, many researchers approach NLU problems using sentence embeddings \cite{reimers-gurevych-2019-sentence}. However, most popular pre-trained models are domain-specific and do not translate well to other domains. This encourages us to explore techniques that maximize the intra-class sentence similarity scores and minimize the inter-class ones. Utilizing deep neural network with a distance metric to learn the feature embedding has been successfully applied to many tasks, such as face recognition \cite{7298682} and speech recognition \cite{Zhang2017}.

For mobile devices, intent detection forms the backbone of Natural Language Understanding (NLU) modules, which can either be used in single-domain or multi-domain conversations \cite{Rastogi_Zang_Sunkara_Gupta_Khaitan_2020}.
We intend to perform multi-domain intent detection for on-device dialog systems, which drives our choice of datasets for experiments. Desai et al. \cite{desai2020lightweight} propose lightweight convolutional representations for on-device task-oriented systems, related to intent classification and other NLP tasks. But they do not benchmark against other pre-trained language models and solely evaluate on a manually curated dataset. In contrast, we compare the efficiency of our proposed model against strong baselines -- including BERT \cite{devlin2019bert} and MobileBERT \cite{sun-etal-2020-mobilebert} on the SNIPS \cite{coucke2018snips} and ATIS \cite{price-1990-evaluation} datasets.

Thus, existing work has focused on simple-but-low-accuracy statistical models or high-accuracy-but-heavy deep learning models. In contrast, we propose a lightweight DNN model that performs at SOTA-competitive accuracy with much less latency and memory footprint than SOTA on edge devices. Furthermore, instead of using pre-trained sentence level features, we employ a two-phase training approach, wherein we train a sentence encoder using a Deep Siamese Network for our multi-domain intent detection.

\section{Methodology}\label{sec:methodology}

In this section, we describe LIDSNet architecture and its two phase training approach. It consists of a sentence encoder that acts as a feature extractor and utilizes character and word features. In the phase I training, the sentence encoder is trained using Siamese Network with Triplet Loss and is utilized to detect the intent of the utterance in training phase II.

\subsection{Data Representation Techniques}\label{sec:dataRepresentationTechniques}

Combining character and word level input representations has shown great success in NLP domain. This is because word representation is suitable for relation classification but it does not perform well on short, informal, and conversational texts, whereas character representation effectively handles misspelt and Out-of-Vocabulary (OOV) words. To leverage the best of both representations, our proposed architecture employs a combination of both.

\subsubsection{Character level features}\label{sec:characterLevelFeatures}

We model morphology by incorporating character level representations of words. This boosts the accuracy of neural  models by learning rich semantic and orthographic features. 
The char-level CNN technique utilized in our model encodes the input character, $c_i$, into $e_{c_i}$. These encoded vectors are then passed through a 1D convolution layer followed by max-pooling. Our model has two such CNN blocks with different kernel sizes to capture multiple character level $n$-gram features, which are then concatenated as defined in \eqref{eq:concatWordChar}.

\subsubsection{Fine-tuned Word Vectors}\label{sec:fineTunedWordVectors}

We hypothesize that using knowledge from pre-trained word embeddings can enable an improved understanding of semantics and inter-word relationships over random weights initialization. We are utilizing language semantic knowledge acquired from the pre-trained embeddings and then fine-tuning it for our task.

We dynamically fine-tune word embedding, $e_{w_j}$, for each word, $w_j$, in the training vocabulary. This word embedding is then concatenated with the character level representation of corresponding word to form an output embedding, $o_{w_j}$:
\begin{multline}
	o_{w_j} = \text{concat}\big(e_{w_j}, \text{CNN}_1(e_{c_1}, ..., e_{c_i}, ..., e_{c_n}), \\ \text{CNN}_2(e_{c_1}, ..., e_{c_i}, ..., e_{c_n})\big)
	\label{eq:concatWordChar}
\end{multline}

\subsection{Proposed Architecture}\label{sec:proposedArchitecture}

\begin{figure}[t]
	\centering
	\includegraphics[width=0.9\linewidth]{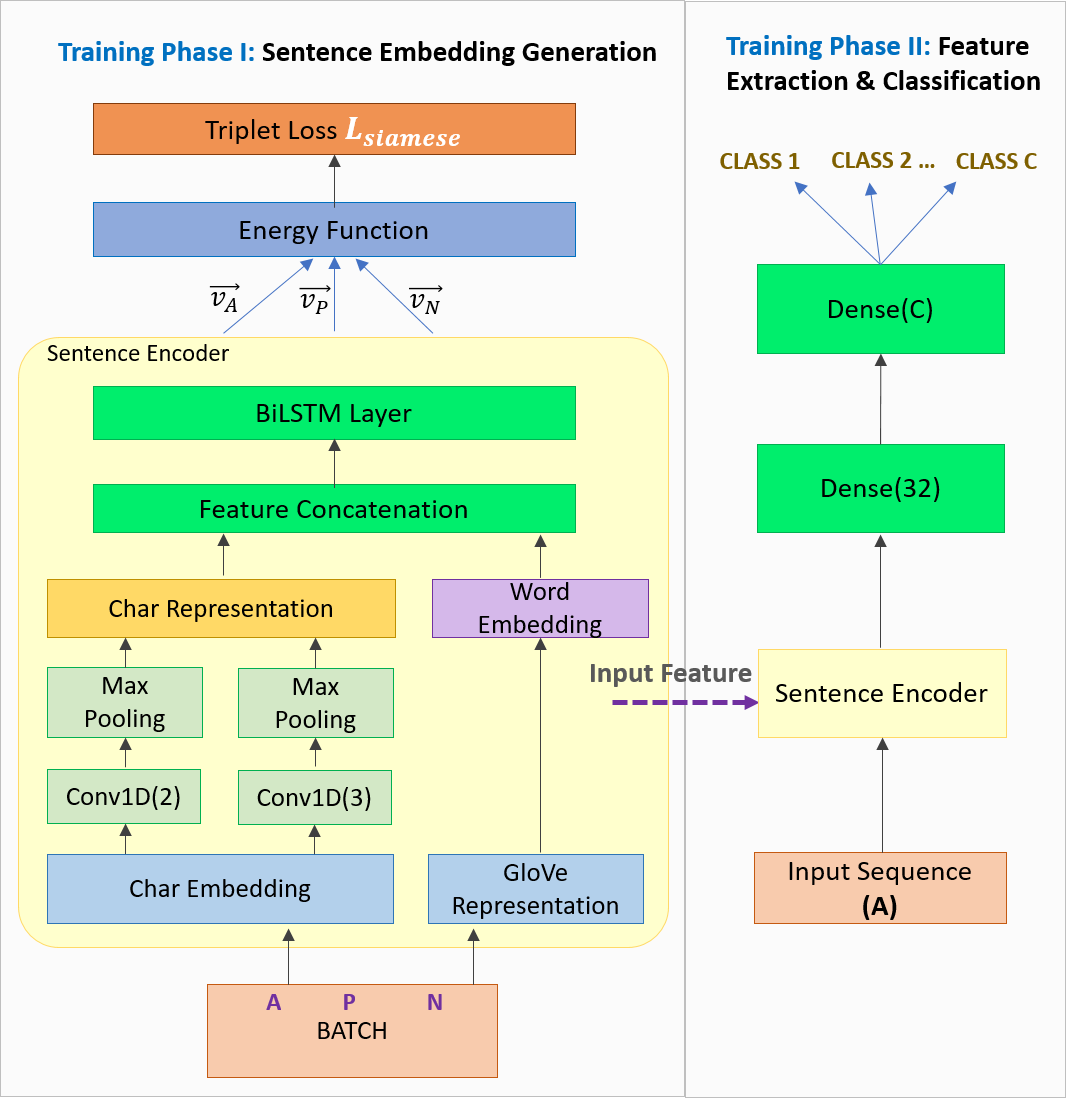}
	\caption{Proposed LIDSNet Architecture}
	\label{fig:proposedArchitecture}
\end{figure}

Siamese Network \cite{Chicco2021} with Triplet Loss \cite{7298682} consists of three identical neural sub-networks with shared weights. Our proposed model consists of a sentence encoder that is trained in two phases. We use two blocks of CNN with different filter sizes to encode multiple character level features as illustrated in Fig.~\ref{fig:proposedArchitecture}. This helps in obtaining representations of rare words \cite{bojanowski-etal-2017-enriching} and modeling sub-word structures. We apply transfer learning by using a subset of pre-trained word embeddings that capture semantics. These embeddings are concatenated with character level features of the corresponding words. The resultant word level features are then fed to a BiLSTM layer to obtain sentence level representations. In training phase II, we pass these through a Feed Forward Neural Network (FFNN), which acts as a classifier to compute the probability distribution over the defined intents.

\subsection{Training Phase I}\label{sec:trainingPhaseI}

In the first phase, we train the Siamese Network with the Triplet Loss function. We use standard train splits of the entire dataset and do not include any additional samples for model training. Each input batch sample consists of three text sequences. Two of these, which we refer to as $A$ (Anchor) and $P$ (Positive), are from the same intent class, while the third sequence, $N$ (Negative), belongs to a different class. For each intent class, we generate all possible $<A,P>$ combinations and sample 50k pairs from these. For each $<A,P>$ pair, we randomly select an $N$-sequence.

Let the shared parameters that need to be optimized be $W$. Our objective is to tune the weights of the sentence encoder in such a way that the sentence representations for $A$ and $P$ should be closer in vector space than $A$ and $N$. Mathematically, our Triplet Loss is computed as:
\begin{multline}
	L_{\text{siamese}}\left(W, \left(A,P,N\right) \right) = \sum_{i=1}^{m} \max \Big( 0, \alpha - s\left(\overrightarrow{v_A}, \overrightarrow{v_P}\right) \\ + s\left(\overrightarrow{v_A}, \overrightarrow{v_N}\right) \Big)
	\label{eq:l_siamese}
\end{multline}

where, $\alpha$ is a hyperparameter to control the margin between positive and negative inputs, $m$ is the total number of training samples, and $s\left(\overrightarrow{v_x}, \overrightarrow{v_y}\right)$ is the cosine similarity between two sequence representations, $\overrightarrow{v_x}$ and  $\overrightarrow{v_y}$, given by $\frac{\overrightarrow{v_x} \cdot \overrightarrow{v_y}}{\left|\left|\overrightarrow{v_x}\right|\right|\left|\left|\overrightarrow{v_y}\right|\right|}$.\\

The Triplet Loss tries to maximize the similarity between sentence representations of $A$ and $P$, while minimizing the similarity between those of $A$ and $N$. With this training phase, our aim is to tune the weights of the shared neural network in a way that it is able to segregate and create a distinction between the sentence embeddings of positive (same intent) and negative (different intent) training sequences. After training, the same sentence encoder is used as a feature extractor for the Classifier network in training phase II.

\subsection{Training Phase II}\label{sec:trainingPhaseII}

This training phase focuses on identifying the most probable intent class for the input sequence. The sentence representations from phase I are passed through two dense layers with Softmax activation function to emit probability scores for each intent. We use categorical cross-entropy loss function, $L_{\text{classifier}}$, defined as $- \sum_{i=1}^{c} y_i \cdot \log \widehat{y_i}$, where, $c$ is the output size, $y_i$ represents the actual label of input $i$, and $\widehat{y_i}$ represents the prediction of our model.

\section{Experimental Setup}\label{sec:experimentalSetup}

\subsection{Datasets}\label{sec:datasets}

The experiments presented in this paper are carried out on our Custom dataset (for conversational texts) along with SNIPS \cite{coucke2018snips} and ATIS \cite{price-1990-evaluation} public datasets.

\subsubsection{Custom Dataset}\label{sec:customDataset}

\begin{figure}[t]
	\centering
	\includegraphics[width=0.6\linewidth]{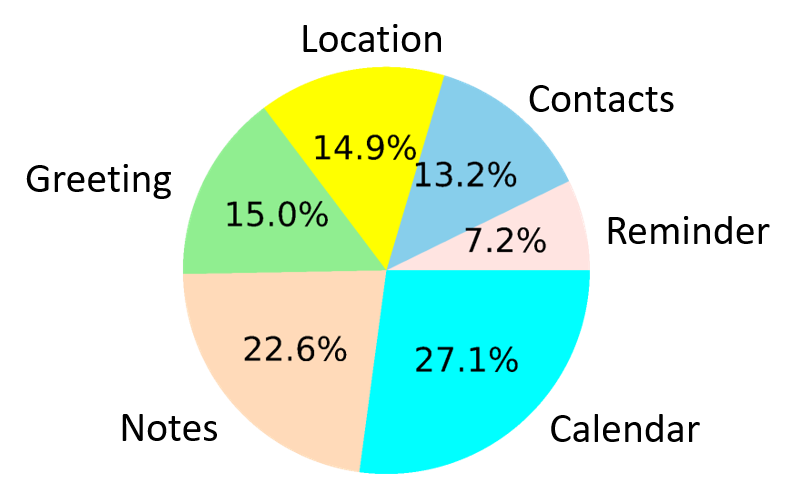}
	\caption{Class distribution in Custom Dataset}
	\label{fig:customDatasetClassDistribution}
\end{figure}

For the task of user intent understanding for conversational texts, we curate our own dataset using CLINIC150 \cite{larson-etal-2019-evaluation} and HWU64 \cite{Liu2019BenchmarkingNL} datasets. We map the relevant intent labels from these public datasets to our intent classes as shown in Table~\ref{tab:customDatasetIntents}. Using this approach, we extract 4148 samples mapping to our six intents. This dataset is then split into 90\% training set and 10\% validation set. The intent-wise distribution of this data is illustrated in Fig.~\ref{fig:customDatasetClassDistribution}. For performance evaluation, we prepare an unseen test set of 298 samples by crowdsourcing conversational texts.

\begin{table}[t]
	\caption{Custom Dataset Intents}
	\centering
	\resizebox{\linewidth}{!}{%
		\begin{tabular}{c c c c c}
			\toprule
			\textbf{Intents from}         & \multirow{2}{*}{+} & \textbf{Intents from}                     & \multirow{2}{*}{=} & \textbf{Unified}                    \\
			\textbf{CLINIC150}           &                    & \textbf{HWU64}                            &                    & \textbf{intents}                    \\ \midrule
			{\scriptsize shopping\_list,}     &                    &                                           &                    & \multirow{4}{*}{Notes}              \\
			{\scriptsize shopping\_list\_update,} &                    & lists\_createoradd,                       &                    &                                     \\
			{\scriptsize todo\_list,}       &                    & lists\_query                              &                    &                                     \\
			{\scriptsize todo\_list\_update}    &                    &                                           &                    & \Bstrut                             \\ \hline
			{\scriptsize calendar,}        &                    &                                           &                    & \multirow{4}{*}{Calendar}   \Tstrut \\
			{\scriptsize meeting\_schedule, }   &                    & calendar\_query,                          &                    &                                     \\
			{\scriptsize calendar\_update, }    &                    & calendar\_remove,                         &                    &                                     \\
			{\scriptsize schedule\_meeting }    &                    & calendar\_set                             &                    & \Bstrut                             \\ \hline
			{\scriptsize reminder,   }       &                    & \multirow{2}{*}{$-$}                      &                    & \multirow{2}{*}{Reminder}   \Tstrut \\
			{\scriptsize  reminder\_update }    &                    &                                           &                    & \Bstrut                             \\ \hline
			{\scriptsize make\_call, }       &                    & email\_addcontact,                        &                    & \multirow{2}{*}{Contacts}   \Tstrut \\
			{\scriptsize text}           &                    & email\_querycontact                       &                    & \Bstrut                             \\ \hline
			{\scriptsize directions,}       &                    & \multirow{3}{*}{recommendation\_location} &                    & \multirow{3}{*}{Location}   \Tstrut \\
			{\scriptsize current\_location,}    &                    &                                           &                    &                                     \\
			{\scriptsize share\_location}     &                    &                                           &                    & \Bstrut                             \\ \hline
			{\scriptsize greeting,   }       &                    & \multirow{3}{*}{general\_praise}          &                    & \multirow{3}{*}{Greeting}   \Tstrut \\
			{\scriptsize goodbye,    }       &                    &                                           &                    &                                     \\
			{\scriptsize thank\_you  }       &                    &                                           &                    & \Bstrut                             \\ \bottomrule
		\end{tabular}%
	}
	\label{tab:customDatasetIntents}
\end{table}

\begin{table}[b]
	\caption{Dataset Statistics}
	\centering
	\begin{tabular}{c r r r}
		\toprule
		\multirow{2}{*}{\textbf{Dataset}} & \multirow{2}{*}{\textbf{SNIPS}} & \multirow{2}{*}{\textbf{ATIS}} & \textbf{Custom}                \\
		&                                 &                                & \textbf{Dataset}               \\ \midrule
		Training Data           & 13084                           & 4478                           & 3734                           \\
		Validation Data          & 700                             & 500                            & 414             \Tstrut        \\
		Test Data             & 700                             & 893                            & 298             \Tstrut        \\
		Vocabulary Size          & 11241                           & 722                            & 2088            \Tstrut        \\
		\# of Intents           & 7                               & 21                             & 6               \Tstrut\Bstrut \\ \bottomrule
	\end{tabular}
	\label{tab:datasetStatistics}
\end{table}

\subsubsection{Public Datasets}\label{sec:publicDatasets}

To evaluate the efficiency of our proposed model, we also evaluate LIDSNet on two real-world datasets, widely used to benchmark intent detection models. The first are the custom-intent engines collected by SNIPS \cite{coucke2018snips}, and the second is ATIS \cite{price-1990-evaluation} dataset, containing audio recordings of airline travel information. Table~\ref{tab:datasetStatistics} presents statistics of all three datasets used in our experiments. We use the same train-validation-test distribution for pre-processed public datasets as Goo et al. \cite{goo-etal-2018-slot}.

\subsection{Implementation Details}\label{sec:implementationDetails}

In sentence encoder, we use Conv1D of filter sizes 2 and 3. The filter count is set to 16. We use a subset of the 50-dim GloVe \cite{pennington-etal-2014-glove} embeddings corresponding to the training set vocabulary. We apply regular and recurrent dropouts with value 0.2. The batch size is set to 512 and 32 for phases I and II respectively. A margin of 0.2 and 24 hidden units in LSTM give the best results as discussed in subsection \ref{sec:ablationStudy}. A constant learning rate of 0.001 is used with Adam \cite{DBLP:journals/corr/KingmaB14} optimizer for training the model in both phases. We use same set of hyperparameters to build model for all three datasets. We choose accuracy as the evaluation metric as it is commonly used by the existing models. We build all our models using TensorFlow framework. Furthermore, we convert them to TensorFlow Lite format for deploying these models on mobile and edge devices.

\section{Evaluation Results}\label{sec:evaluationResults}

We perform an ablation study of the different aspects of LIDSNet in order to show the impact of every component of its architecture. Furthermore, we analyze the impact of varying hyperparameters on performance. We also benchmark our model against SOTA intent detection models and compare its computational efficiency with baseline BERT-based models.

\subsection{Ablation Study}\label{sec:ablationStudy}

We investigate the effect of different architectural and methodical choices, the results of which are presented in Table~\ref{tab:ablationStudy}. We train and evaluate our final model against other model variants. These models are named as $i$\verb|P_embeddingType|, where $i=1$ represents that model is trained only for classification task (bypassing phase 1), and $i=2$ implies that model is trained for both phases.

\begin{table}[t]
	\caption{Ablation Study}
	\centering
	\begin{tabular}{l c c c}
		\toprule
		\multirow{3}{*}{\textbf{Model}} &                               \multicolumn{3}{c}{\textbf{Accuracy (\%)}}                                \\ \cline{2-4}
		                                & \multirow{2}{*}{\textbf{SNIPS}} & \multirow{2}{*}{\textbf{ATIS}} & \textbf{Custom}  \Tstrut             \\
		                                &                                 &                                & \textbf{Dataset}                     \\ \midrule
		\verb|1P_random| (base)         & 96.43                           & 94.85                          & 90.60                  \Tstrut       \\
		\verb|1P_GloVe|                 & 96.71                           & 95.41                          & 91.61                  \Tstrut       \\
		\verb|2P_Char|                  & 96.71                           & 95.30                          & 90.94                  \Tstrut       \\
		\verb|2P_random|                & 96.86                           & 95.52                          & 91.95                  \Tstrut       \\
		\verb|2P_fastText|              & 96.86                           & 95.63                          & 91.61                  \Tstrut       \\
		\verb|2P_GloVe| (LIDSNet)       & \textbf{98.00}                  & \textbf{95.97}                 & \textbf{93.62}	       \Tstrut\Bstrut \\ \bottomrule
	\end{tabular}
	\label{tab:ablationStudy}
\end{table}

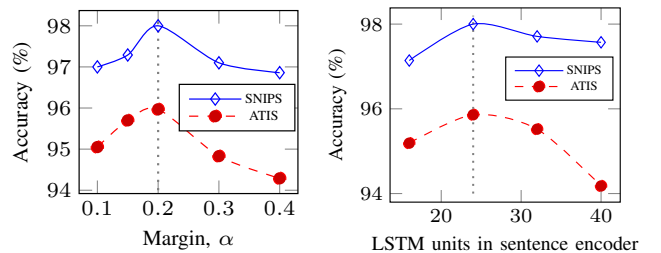
\begin{figure}[b]
	\centering
	\begin{minipage}[t]{0.462\linewidth}
		\resizebox{\linewidth}{!}{
			\pgfplotsset{
				width=\linewidth
			}
			\begin{tikzpicture}
				\scriptsize 
				\begin{axis}[xlabel={Margin, $\alpha$},ylabel={Accuracy (\%)},
					xticklabel style={/pgf/number format/fixed},
					legend style ={ at={(0.455,0.6)}, 
						anchor=north west, draw=black, 
						fill=white,align=left,
						nodes={scale=0.5, transform shape}
					},
					smooth,
					width=\linewidth,
					height=0.9\linewidth
					]
					
					\draw[line width=0.25mm, dotted, color=gray] (0.2, 94) -- (0.2, 98.2);
					
					\addplot+[smooth, mark=diamond] coordinates{						
						(0.1, 97)
						(0.15, 97.29)
						(0.2, 98)
						(0.3, 97.1)
						(0.4, 96.86)
					};
					\addlegendentry{\small SNIPS};
					
					\addplot+[dashed, mark=*] coordinates{
						(0.1, 95.05)
						(0.15, 95.7)
						(0.2, 95.97)
						(0.3, 94.83)
						(0.4, 94.29)
					};
					\addlegendentry{\small ATIS};
					
				\end{axis}
			\end{tikzpicture}
		}
		\captionsetup{labelformat=empty}
		\label{fig:hyperparameterAnalysis_marginVsAccuracy}
	\end{minipage}
	\begin{minipage}[t]{0.5\linewidth}
		\resizebox{\linewidth}{!}{
			\pgfplotsset{
				width=\linewidth
			}
			\begin{tikzpicture}
				\scriptsize 
				\begin{axis}[xlabel={LSTM units in sentence encoder},ylabel={Accuracy (\%)},
					xticklabel style={/pgf/number format/fixed},
					legend style ={ at={(0.5,0.75)}, 
						anchor=north west, draw=black, 
						fill=white,align=left,
						nodes={scale=0.5, transform shape}
					},
					smooth,
					width=\linewidth,
					height=0.9\linewidth
					]
					
					\draw[line width=0.25mm, dotted, color=gray] (24, 94) -- (24, 98.2);
					
					\addplot+[smooth, mark=diamond] coordinates{						
						(16, 97.14)
						(24, 98)
						(32, 97.71)
						(40, 97.57)
					};
					\addlegendentry{\small SNIPS};
					
					\addplot+[dashed, mark=*] coordinates{
						(16, 95.19)
						(24, 95.86)
						(32, 95.52)
						(40, 94.18)
					};
					\addlegendentry{\small ATIS};
					
				\end{axis}
			\end{tikzpicture}
		}
		\captionsetup{labelformat=empty}
		\label{fig:hyperparameterAnalysis_LSTMUnitsVsAccuracy}
	\end{minipage}
	\caption{Hyperparameter Analysis}
	\label{fig:hyperparameterAnalysis}
\end{figure}

\begin{figure*}
	\centering
	\begin{minipage}[t]{0.4\textwidth}
		\includegraphics[width=0.86\linewidth]{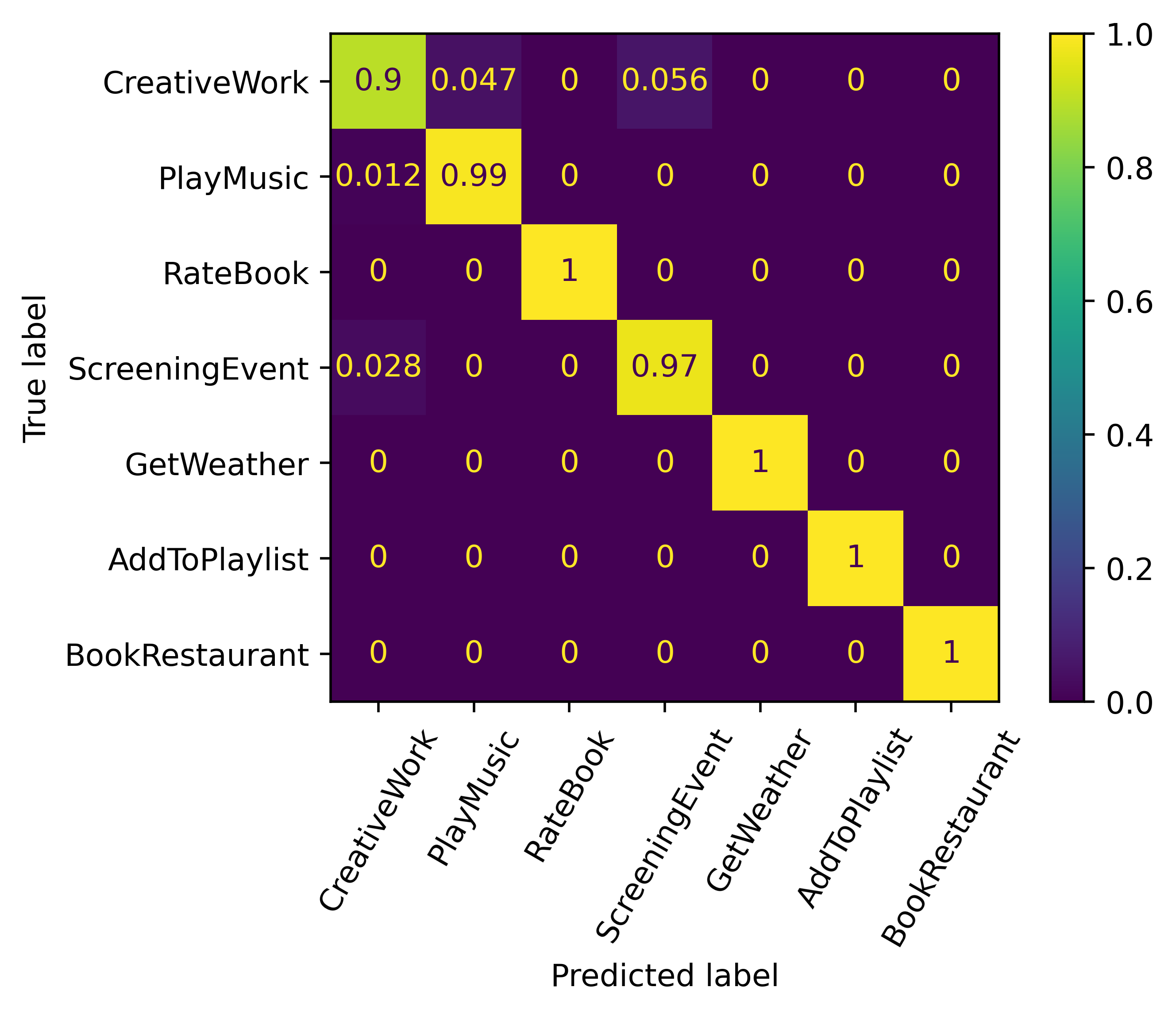}
		\captionsetup{labelformat=empty}
		\caption*{(a) on SNIPS dataset}
		\label{fig:confusionMatrixSNIPSDataset}
	\end{minipage}
	\quad
	\begin{minipage}[t]{0.4\textwidth}
		\includegraphics[width=0.86\linewidth]{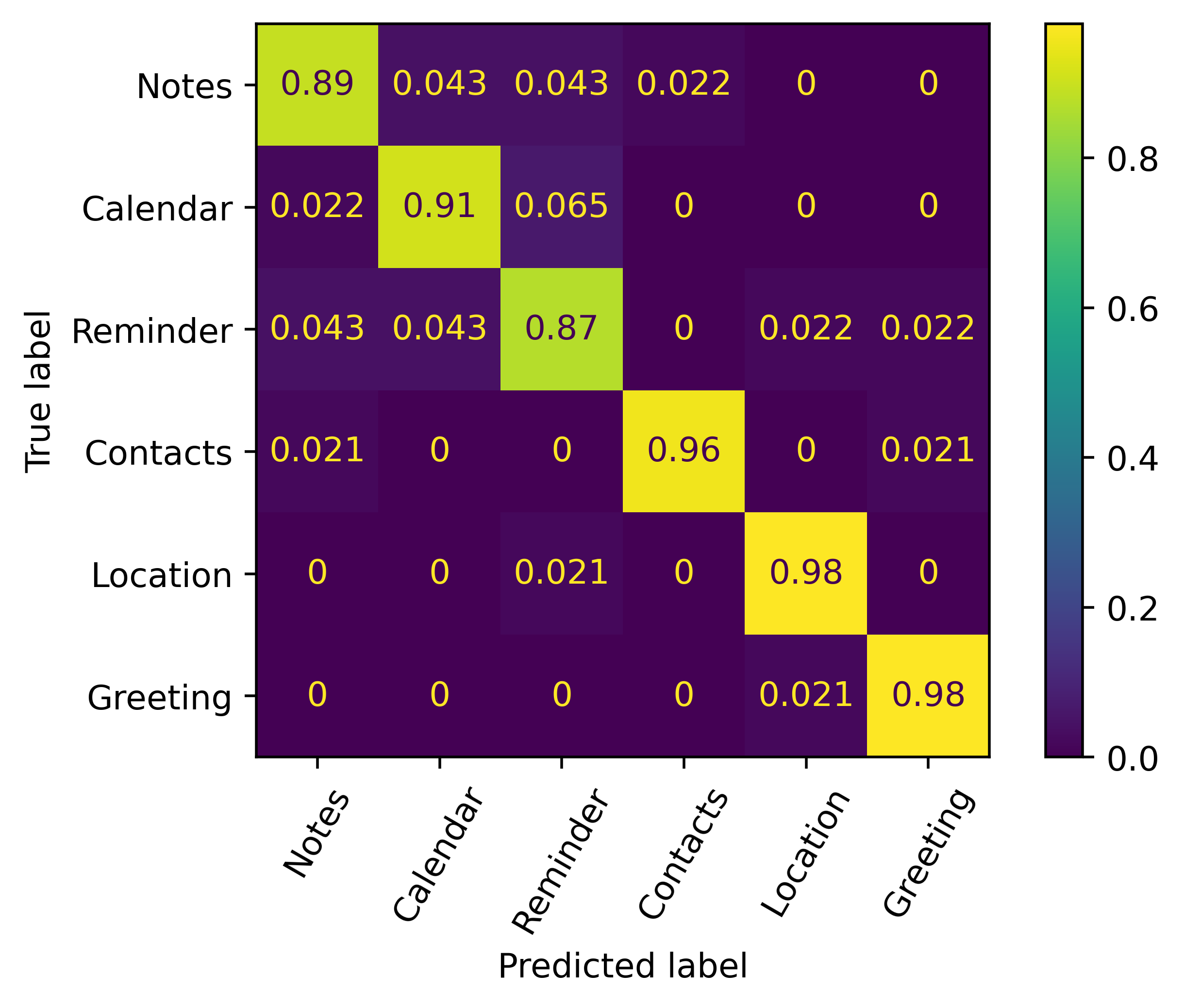}
		\captionsetup{labelformat=empty}
		\caption*{(b) on Custom dataset}
		\label{fig:confusionMatrixCustomDataset}
	\end{minipage}
	\caption{Confusion Matrices summarizing the Classification Performance of LIDSNet on SNIPS and Custom test data}
	\label{fig:confusionMatrices}
\end{figure*}

\begin{enumerate}
	\item \verb|1P_random| (base classifier): Model is trained with randomly initialized word embeddings, bypassing phase I training. Comparison of LIDSNet with base classifier shows the combined effect of two phase training methodology and initialization of word vectors with pre-trained embeddings in improving performance.
	
	\item \verb|1P_GloVe|: Model is trained with GloVe embeddings initialization, bypassing phase I training. We use this to investigate the effectiveness of knowledge gained in phase I training at improving classification performance.
	
	\item \verb|2P_Char|: Only character-level embeddings are utilized in sentence encoder and is trained for both phases. This helps highlight the importance of semantic and syntactic knowledge gained through word embeddings.
	
	\item \verb|2P_random|: Random word embedding initialization and model is trained for both phases. This variant helps us investigate the improvement due to knowledge transfer of pre-trained word embeddings.
	
	\item \verb|2P_fastText|: Embedding weights are initialized from fastText \cite{bojanowski-etal-2017-enriching} and model is trained for both phases. 
	
	\item \verb|2P_GloVe| (LIDSNet): Embeddings weights are initialized from GloVe and model is trained for both phases.
\end{enumerate}

\begin{table}[b]
	\caption{Comparison with SOTA Models}
	\centering
	\resizebox{\columnwidth}{!}{
		\begin{tabular}{c l l r}
			\toprule
			\multirow{2}{*}{\textbf{Model}}               & \multicolumn{2}{c}{\textbf{Accuracy (\%)}} & \textbf{Model Size}                  \\ \cline{2-3}
			& \textbf{SNIPS} & \textbf{ATIS}             & \textbf{(MB): SNIPS} \Tstrut         \\ \midrule
			Stack-Propagation + BERT \cite{qin2019stackpropagation}   & 99.00          & 97.50                     & $>$1200.00           \Tstrut         \\
			\textbf{LIDSNet}                      & 98.00          & 95.97                     & \textbf{0.63}        \Tstrut         \\
			Stack-Propagation \cite{qin2019stackpropagation}      & 98.00          & 96.90                     & 3.32                  \Tstrut        \\
			Capsule-NLU \cite{zhang2019joint}              & 97.70          & 95.00                     & 643.27                \Tstrut        \\
			SF-ID (BLSTM) network \cite{e2019novel}           & 97.43          & 97.76                     & 11.61                 \Tstrut        \\
			Slot-Gated BiLSTM with Attention \cite{goo-etal-2018-slot} & 97.00          & 94.10                     & 11.57                 \Tstrut\Bstrut \\ \bottomrule
		\end{tabular}
	}
	\label{tab:comparisonWithSOTAModels}
\end{table}

We observe that LIDSNet achieves absolute accuracy improvement of 1.29\% and 0.56\% on SNIPS and ATIS respectively when compared to \verb|1P_GloVe| classifier. This empirically proves that our two phase training methodology is useful. Moreover, by comparing the accuracy of \verb|2P_Char| and LIDSNet, we show the effectiveness of combining word and character level features in learning a better representation for classification. The impact of GloVe embeddings can be seen from the fact that LIDSNet achieves 1.14\% and  0.45\% improvement over \verb|2P_random| classifier on SNIPS and ATIS respectively. With LIDSNet, we obtain an accuracy improvement of 1.57\% and 1.12\% on SNIPS and ATIS respectively, compared to the base classifier. This improvement shows the combined effect of our methodical choices. The model sizes of LIDSNet, when trained on SNIPS, ATIS, and Custom datasets, are only 0.63 MB, 0.12 MB and 0.19 MB respectively. Fig.~\ref{fig:hyperparameterAnalysis} illustrates the effect of varying hyperparameters on model accuracy. The classification performance of LIDSNet is presented in Fig. \ref{fig:confusionMatrices}.

\subsection{Comparison with SOTA}\label{sec:comparisonWithSOTA}

We compare our best model with other SOTA methods on SNIPS and ATIS datasets. Table \ref{tab:comparisonWithSOTAModels} shows that LIDSNet beats most SOTA models in terms of accuracy even with the lowest memory footprint. For model size comparison, we re-implemented the models and obtained the results on same datasets. In SNIPS dataset, we achieve the second highest accuracy of 98.00\% (next only to Stack-Propagation with BERT \cite{qin2019stackpropagation}) with the lowest model size of 0.63 MB. This can enable effective deployment of LIDSNet on edge devices such as mobile phones. In ATIS dataset, we observe absolute accuracy improvement of 1.87\% and 0.97\% over Slot-Gated BiLSTM with Attention \cite{goo-etal-2018-slot} and Capsule-NLU \cite{zhang2019joint} respectively. However, our accuracy is slightly lower than some baselines for ATIS, but given the tiny model size of LIDSNet, it offers a compelling accuracy-memory trade-off for inference on resource constraint edge devices.

\begin{table}[b]
	\caption{Benchmarking LIDSNet for Mobile Inference}
	\centering
	\resizebox{\columnwidth}{!}{
		\begin{tabular}{c l l c r r}
			\toprule
			            \multirow{2}{*}{\textbf{Model}}              &    \multicolumn{2}{c}{\textbf{Accuracy (\%)}}     & \textbf{Params} & \textbf{Latency}     & \multirow{2}{*}{\textbf{Speedup}}              \\ \cline{2-3}
			                                                         & \textbf{SNIPS}          & \textbf{ATIS}           & \textbf{(M)}    & \textbf{(ms)}        & \Tstrut                                        \\ \midrule
			     BERT\textsubscript{BASE} \cite{devlin2019bert}      & 98.26                   & 97.16                   & 110             & 1580                 & 1.0x                            \Tstrut\Bstrut \\ \hline
			          DistilBERT \cite{sanh2020distilbert}           & 97.94                   & 96.98                   & 66              & 781                  & 2.0x                            \Tstrut\Bstrut \\ \hline
			       MobileBERT \cite{sun-etal-2020-mobilebert}        & 97.71                   & 96.30                   & 24.6            & 545                  & 2.9x                            \Tstrut\Bstrut \\ \hline
			TinyBERT\textsubscript{4} \cite{jiao-etal-2020-tinybert} & 97.43                   & 95.97                   & 14.5            & 162                  & 9.8x                            \Tstrut\Bstrut \\ \hline
			                                                         &                         &                         & 0.59 (SNIPS)    &                      & \Tstrut                         \Tstrut        \\
			           \multirow{-2}{*}{\textbf{LIDSNet}}            & \multirow{-2}{*}{98.00} & \multirow{-2}{*}{95.97} & 0.065 (ATIS)    & \multirow{-2}{*}{18} & \multirow{-2}{*}{\textbf{87.0x}}  \Bstrut      \\ \bottomrule
		\end{tabular}
	}
	\label{tab:mobileInferenceBenchmarking}
\end{table}

\subsection{Computational Experiments}\label{sec:computationalExperiments}

To understand how our LIDSNet model performs in the absence of significant computational resources, we benchmark its latency against fine-tuned BERTs. Model inferencing experiments are conducted on Samsung Galaxy S20 device (8 GB RAM, 2 GHz octa-core Exynos 990 processor).

Model parameters of LIDSNet on SNIPS and ATIS are 0.59M and 0.065M respectively. The higher number of parameters with SNIPS is due to its large vocabulary. Table~\ref{tab:mobileInferenceBenchmarking} shows that BERT\textsubscript{BASE} \cite{devlin2019bert} performs best with 98.26\% and 97.16\% accuracy on SNIPS and ATIS respectively. However, LIDSNet stands out on top with 0.53\% parameters and 87x inference speedup in comparison to BERT\textsubscript{BASE}. Compared to 4-layer TinyBERT\textsubscript{4} \cite{jiao-etal-2020-tinybert}, LIDSNet is 24x smaller and yet 9x faster. The results also show that our proposed model is 41x smaller, 30x faster than MobileBERT \cite{sun-etal-2020-mobilebert}, and 111x smaller, 43x faster than DistilBERT \cite{sanh2020distilbert}. Our model has a maximum inference time of only 18 milliseconds as reported in Table~\ref{tab:mobileInferenceBenchmarking}. All above memory-latency comparative analysis is with respect to SNIPS training. Since these BERT-based models have an excess of tens of millions of parameters, they are impractical to be deployed on-device, where our model significantly outperforms all the baselines.

\section{Conclusion}\label{sec:conclusion}

In this paper, we propose a lightweight, fast, and accurate LIDSNet model for intent classification. We adopt a two phase training methodology and empirically demonstrate the advantage of transfer learning with fine-tuned vectors in improving performance. Our experimental analysis on SNIPS (98.00\%, 0.63 MB), ATIS (95.97\%, 0.12 MB), and custom dataset (93.62\%, 0.19 MB) proves that LIDSNet achieves SOTA-competitive accuracy with the lowest memory footprint. Furthermore, we explore and analyze how LIDSNet clearly outperforms fine-tuned BERTs in terms of system-specific metrics like ROM and latency which is crucial for creating a commercial conversational AI solution. In the future, we plan to develop a joint model for intent detection and slot filling.

\bibliographystyle{IEEEtran}
\bibliography{bibliography}

\end{document}